# General vs Domain-Specific CNNs: Understanding Pretraining Effects on Brain MRI Tumor Classification

**Helia Abedini[1,*], Saba Rahimi[2], and Reza Vaziri[3]**

1. Department of Computer and Information Technology, Islamic Azad University, Central Tehran Branch, Tehran, Iran; he.abedini@iau.ir, abedinia054@gmail.com
2. Faculty of Pharmacy, Charles University, Prague, Czech Republic; Rahimis@faf.cuni.cz
3. Department of Computer and Information Technology, Islamic Azad University, Central Tehran Branch, Tehran, Iran; Rez.vaziri@iauctb.ac.ir

* Correspondance: abedinia054@gmail.com; Tel: +98-919-767-5691

**Abstract**

The accurate identification of brain tumors from magnetic resonance imaging (MRI) is essential for timely diagnosis and effective therapeutic intervention. While deep convolutional neural networks (CNNs), particularly those pre-trained on extensive datasets, have shown considerable promise in medical image analysis, a key question arises when working with limited data: do models pre-trained on specialized medical image repositories outperform those pre-trained on diverse, general-domain datasets?

This research presents a comparative analysis of three distinct pre-trained CNN architectures for brain tumor classification: RadImageNet DenseNet121, which leverages pre-training on medical-domain data, alongside two modern general-purpose networks, EfficientNetV2S and ConvNeXt-Tiny. All models were trained and fine-tuned under uniform experimental conditions using a modestly sized brain MRI dataset to maintain consistency in evaluation. The experimental outcomes indicate that ConvNeXt-Tiny delivered the best performance, achieving 93% test accuracy, followed by EfficientNetV2S at 85%. In contrast, RadImageNet DenseNet121 attained only 68% accuracy and exhibited higher loss, indicating limited generalization capability despite its domain-specific pre-training.

These observations imply that pre-training on medical-domain data does not necessarily guarantee superior performance in data-scarce scenarios. Conversely, contemporary general-purpose CNNs with deeper architectures, pre-trained on large-scale diverse datasets, may offer more effective transfer learning for specialized diagnostic tasks in medical imaging.

---

## 1. Introduction

Magnetic resonance imaging (MRI) is a core element of contemporary neuro-oncology, playing a major role in the early diagnosis and treatment of brain tumors [1]. Deep convolutional neural networks (CNNs) have changed the way computer-aided diagnosis works by the automation of feature extraction and by the obtaining of high accuracy in various medical imaging tasks [2, 3]. However, their performance and generalization ability are often constrained by the lack of annotated medical data, particularly in specialized fields such as neuroimaging, where small datasets limit the effectiveness of pretrained models.

Recent studies have explored the use of pretrained CNN architectures for medical image analysis, leveraging large, general-purpose datasets such as ImageNet. These models, pretrained on diverse images, often serve as strong feature extractors for medical tasks. On the other hand, domain-specific models, pretrained on medical datasets, are expected to perform better by adapting to the unique characteristics of medical images. Despite this, the effectiveness of domain-specific pretrained models on small datasets remains unclear, and there is limited research comparing these models against general-purpose models in such contexts.

This study systematically evaluates the performance of three pretrained CNN models for brain tumor classification: RadImageNet DenseNet121 (trained on a medical dataset), EfficientNetV2S, and ConvNeXt-Tiny (modern general-purpose models pretrained on ImageNet). The models are trained and fine-tuned under identical conditions using a limited-size brain MRI dataset to ensure a fair comparison. The primary objective is to determine which type of pretrained model, whether domain-specific or general-purpose, performs better on a small dataset, specifically in terms of accuracy and generalization to unseen data.

The outcome of this study has significant implications for medical image analysis, especially in scenarios where annotated data is scarce. If general-purpose models outperform domain-specific ones, it would suggest that pretrained models on large datasets may generalize better to medical tasks, even when data is limited. These findings could lead to better-informed decisions when selecting models for medical applications, particularly in clinical settings where access to large datasets is a challenge.

## 2. Literature Review

### 2.1 Overview of Brain Tumor Detection using MRI

Brain tumor detection through MRI scans is an essential task for early diagnosis, treatment planning, and patient management. MRI scans provide high-resolution images that allow medical professionals to identify and differentiate between various types of brain tumors, such as glioma, meningioma, and pituitary tumors. Traditionally, image analysis for tumor detection has relied on manual methods, which are time-consuming and prone to human error. However, with the advent of deep learning, particularly CNNs, automated methods for classifying brain tumors have shown great promise [4].

### 2.2 Role of Deep Learning in Medical Imaging

Deep convolutional neural networks (CNNs) have emerged as the dominant approach in medical image analysis, outperforming traditional machine learning algorithms in a variety of tasks, including classification, segmentation, and detection. CNNs are particularly well-suited for tasks involving high-dimensional data such as medical images. A critical factor behind the success of CNNs is the ability to learn hierarchical features directly from the data, which makes them powerful in applications where handcrafted features are challenging to define. Several studies have shown that CNNs can surpass human experts in various diagnostic tasks, demonstrating their potential to revolutionize healthcare [5].

### 2.3 Pretrained Models and Transfer Learning in Medical Imaging

The concept of transfer learning, particularly the use of pretrained models, has become central in many medical imaging tasks. Transfer learning involves fine-tuning models that have been pretrained on large-scale general datasets, such as ImageNet, for specific tasks. Pretrained CNN models can be adapted to medical image analysis by utilizing learned features from natural images and then fine-tuning them on domain-specific medical datasets. This approach has been shown to significantly reduce training time and improve performance, especially when labeled medical data is scarce. The ability to transfer learned features across domains has made pretrained models a popular choice in medical image analysis [6].

### 2.4 Domain-Specific Pretraining vs. General-Purpose Pretraining

A key distinction in the use of pretrained models for medical imaging is whether the model has been pretrained on domain-specific data (e.g., medical images) or general-purpose data (e.g., natural images). Models pretrained on medical datasets, such as RadImageNet DenseNet121, are designed to capture features specific to medical images, and are believed to offer advantages in terms of performance for specialized tasks [7].

However, general-purpose pretrained models, like EfficientNetV2S and ConvNeXt-Tiny, trained on large and diverse datasets like ImageNet, may provide a more robust feature extractor that generalizes well across tasks. Research comparing these two approaches, particularly in the context of small datasets, remains limited, and their relative effectiveness is still under debate.

### 2.5 Challenges with Small Datasets in Medical Imaging

One of the main challenges in medical image analysis is the limited availability of annotated data. In many medical domains, including brain tumor detection, acquiring large labeled datasets is expensive and time-consuming. Small datasets pose significant challenges to training deep learning models, as they often lead to overfitting and poor generalization to unseen data. This issue becomes especially relevant when comparing pretrained models, as it is unclear whether domain-specific models trained on small medical datasets will outperform general-purpose models pretrained on large-scale, natural image datasets. Several studies have investigated methods to mitigate these challenges, such as data augmentation and transfer learning, but the effectiveness of these strategies remains context-dependent [8].

### 2.6 Research Gaps and Contribution of the Current Study

While existing research has provided insights into the use of pretrained models for medical image analysis, there remains a lack of consensus on the effectiveness of domain-specific vs general-purpose pretrained models when applied to small datasets. Moreover, most studies have focused on larger datasets, making it unclear how these models would perform under conditions where annotated data is limited. This study aims to address these gaps by systematically evaluating RadImageNet DenseNet121, EfficientNetV2S, and ConvNeXt-Tiny on a small brain tumor MRI dataset, providing a direct comparison of domain-specific and general-purpose pretrained models under small-data conditions.

## 3. Methodology

### 3.1 Overview of the Approach

In this study, we aim to evaluate the performance of different domain-specific and general pretrained models. We implemented three distinct models. For domain-specific model we used RadImageNet DenseNet121, and for the two other general models we implemented EfficientNetV2S and ConvNeXt-Tiny. Each model went through a transfer learning phase, and was fine-tuned with 2 small public datasets of brain tumor MR images from Kaggle, consisting of overall 10287 images in both training and test sets, with identical training and evaluation pipeline, ensuring a fair comparison.

### 3.2 Data Description

The datasets used for this study consists of MRI scans of brain tumors, downloaded from two public Kaggle datasets: Brain Tumor MRI Dataset [9], and Brain Tumor Classification (MRI) [10], both used in training, validation, and testing phase. Both datasets were structured with four different tumor classes: glioma, meningioma, pituitary, and notumor. A summary of the datasets are as follows:

**Brain Tumor MRI Dataset:**

- Training: 5712 images (glioma: 1321, meningioma: 1339, pituitary: 1457, notumor: 1595)
- Testing: 1311 images (glioma: 300, meningioma: 306, pituitary: 300, notumor: 405)

**Brain Tumor Classification (MRI):**

- Training: 2870 images (glioma: 826, meningioma: 822, pituitary: 827, notumor: 395)
- Testing: 394 images (glioma: 100, meningioma: 115, pituitary: 74, notumor: 105)

Each image has a resolution of 512x512 pixels and is in RGB format. We resized the images to 224x224 pixels to match the input size required by the pretrained models.

### 3.3 Model Descriptions

Three CNN architectures were evaluated in this study:

- **RadImageNet DenseNet121**: This model is based on the DenseNet121 architecture pretrained on the RadImageNet dataset, which is a domain-specific medical dataset [11]. The model was loaded without its top classification layer, and the weights were loaded from a pretrained version of DenseNet121 (RadImageNet-DenseNet121_notop.h5). The final classification layer was added, followed by training for the task of brain tumor classification.

- **EfficientNetV2S**: A modern general-purpose CNN model that was pretrained on the large ImageNet dataset [12]. In this study we used the model without its top layer, with the weights initialized from ImageNet, and then fine-tuned for the brain tumor classification task. EfficientNetV2S is known for its model efficiency regarding size and computational resources.

- **ConvNeXt-Tiny**: Similarly, the ConvNeXt-Tiny model, a newly developed architecture that follows the principles of convolutional networks while incorporating ideas from vision transformers, was also pretrained on the ImageNet dataset [13]. Like the EfficientNetV2S, it was used without its top layer, with ImageNet weights, and fine-tuned for the tumor classification task.

### 3.4 Preprocessing and Augmentation

Before training, the MR images underwent preprocessing and augmentation steps to ensure better generalization and robustness:

1. **Preprocessing:**
   Images were resized to 224x224 pixels to match the input size required by pretrained models. For all models, the images were preprocessed using the preprocess_input function from Keras. This function adjusts pixel values (e,g,. scaling, normalizing) to match the preprocessing used during the pretrained model's training.
2. **Data Augmentation**:
   To address the limited size of the dataset and prevent overfitting, a variety of data augmentation techniques were applied during training. These included:
    - Rotation (range of 5 degrees)

- Width and height shifts (5% of the image dimensions)
- Shear range (5%)
- Zoom range (5%)
- Brightness adjustment (90-110% of original brightness)

These augmentations were applied using ImageDataGenerator class from Keras, which ensures that the models see varied range of images during training. This enhances overall generalization of models and prevents overfitting, which happens when the model has not seen enough data and would start memorizing patterns.

Figure I illustrates examples of MRI images after applying the augmentation techniques described above. Each image shows variations in rotation, shift, shear, zoom, and brightness applied during training.

**Figure I:** ***Examples of augmented MRI images.*** *The augmentations include rotation, width/height shifts, shear, zoom, and brightness adjustments. These variations help improve model generalization and prevent overfitting.*

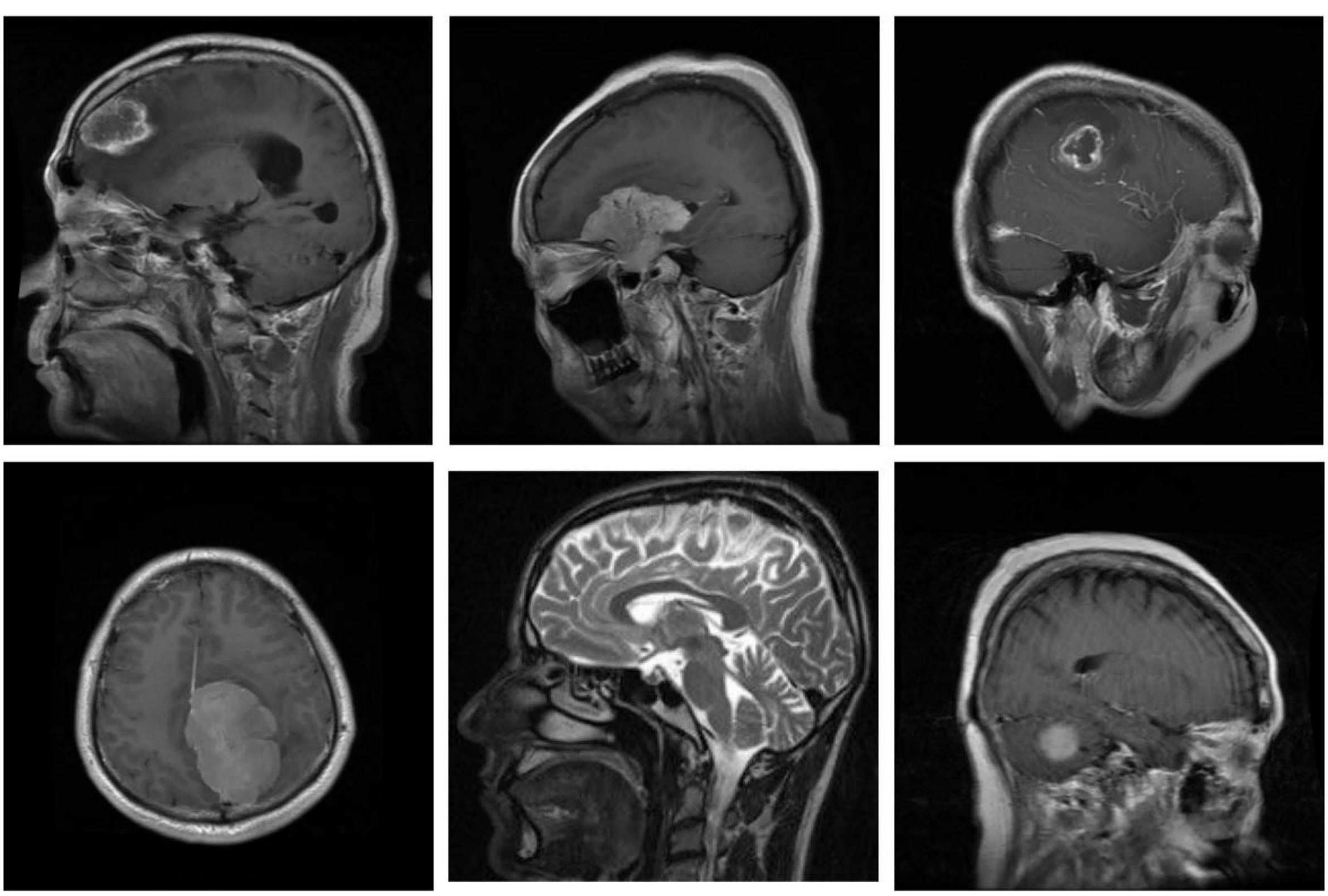

### 3.5 Training Procedure

All three models were trained and fine-tuned under identical conditions to ensure a fair comparison:

- **Data Generators**:
  For the training and validation data, ImageDataGenerator was used to feed batches of images into the model. The training data was split from two training datasets (80% for training, 20% for validation), while two testing datasets were used for final evaluation.

The dataset was processed using the flow_from_directory method, which loads images from the directory structure organized by class.

- **Model Training**:
  Each model was trained in two phases:

  1. **Training Phase 1**: In the first phase, the models were trained with frozen base layers (feature extraction) and only the fully connected layers were trained. This phase used the Adam optimizer with a learning rate of 1e-4 and early stopping based on validation accuracy.

  2. **Fine-Tuning (Phase2)**: In the second phase, the last layers of the pretrained base model were unfrozen, allowing the model to adapt more specifically to the MRI dataset. The learning rate was reduced to 1e-5 during fine-tuning to prevent overfitting.

Models were trained for a maximum of 50 epochs (Phase 1) and 10 epochs (Phase 2), with early stopping to avoid overfitting.

- **Class Weights**:
  Due to the class imbalance in the dataset, class weights were computed using sklearn.utils.class_weight.compute_class_weight. These weights were used during training to give more importance to underrepresented classes.

### 3.6 Evaluation Metrics

To evaluate the performance of the models, the following metrics were used:

- **Accuracy**: The percentage of correctly predicted labels across all test samples.
- **Loss**: The categorical crossentropy loss, which measures the difference between the predicted and true class distributions.
- **Confusion Matrix**: A confusion matrix was generated to visualize how well the models differentiated between the four tumor classes.
- **AUC-ROC**: The Area Under the Receiver Operating Characteristic curve (AUC-ROC) was calculated for each class to assess model performance at various classification thresholds.
- **Precision-Recall Curve**: Precision and recall curves (PR) were plotted to evaluate how well each model balanced the trade-off between false positives and false negatives for each class.

### 3.7 Experimental Setup

All experiments were conducted using TensorFlow 2.x for model development and training. The models were evaluated on both testing datasets, and the results were averaged for a final comparison of model performance.

## 4. Results

In this study, we compared the performances of three pretrained deep learning models: RadImageNet DenseNet121, EfficientNetV2S, and ConvNeXt-Tiny, on a small brain tumor MRI dataset. The models were fine-tuned under identical conditions, while their performance was assessed based on accuracy, loss, and additional evaluation metrics such as confusion matrix, precision-recall curves, and ROC-AUC scores.

### 4.1 Performance Comparison

The models' performances are summarized in Table I. ConvNeXt-Tiny, a general-purpose pretrained model, achieved the highest accuracy of 93% on the test set, outperforming the other models by a significant margin. EfficientNetV2S followed with an accuracy of 85%, while RadImageNet DenseNet121 showed the lowest performance with an accuracy of 68%.

**Table I: *Overall accuracy results of the three pretrained models on the test set.*** *ConvNeXt-Tiny achieved 93% accuracy, followed by EfficientNetV2S at 85%, while RadImageNet DenseNet121 lagged behind at 68%.*

| Model | Overall Test Accuracy |
|---|---|
| **ConvNeXt-Tiny** | 93% |
| **EfficientNetV2S** | 85% |
| **RadImageNet DenseNet121** | 68% |

### 4.2 Confusion Matrix and Per-Class Accuracy

The confusion matrix for each model is provided in Figure II. ConvNeXt-Tiny showed the best overall performance, with high testing accuracy across all classes, followed by EfficientNetV2S. DenseNet121 struggled particularly with classifying certain tumor types, such as meningioma and glioma, resulting in more frequent misclassifications. This is evident in the per-class accuracy bar plot in Figure III, where DenseNet121 showed lower accuracy for some classes.

**Figure II: *Confusion matrices for the three evaluated CNN models.*** *ConvNeXt-Tiny (top-right) demonstrates the most accurate classifications with the highest diagonal values, followed by EfficientNetV2S (top-left). RadImageNet DenseNet121 (bottom) exhibits significant confusion, particularly misclassifying meningioma tumors as glioma or notumor.*

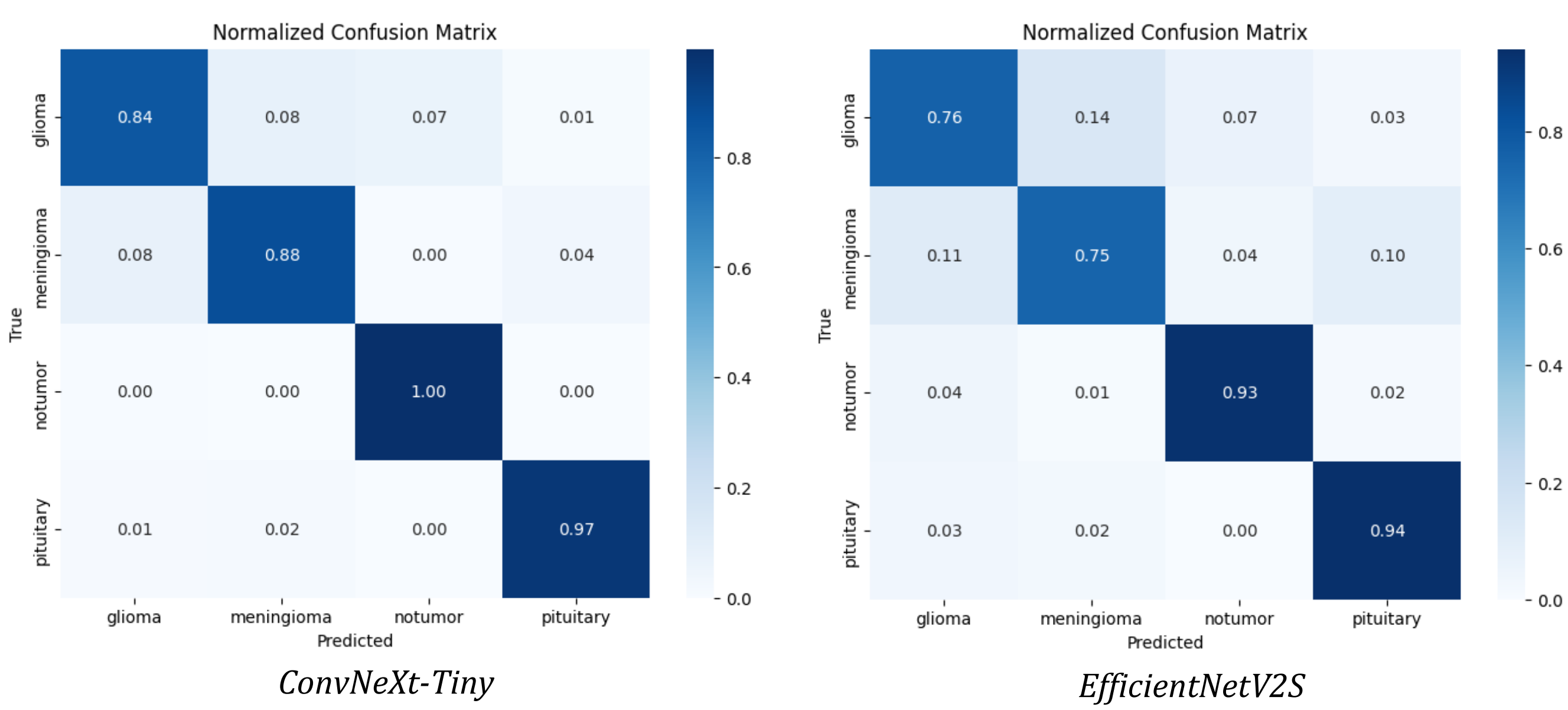

*ConvNeXt-Tiny*

*EfficientNetV2S*

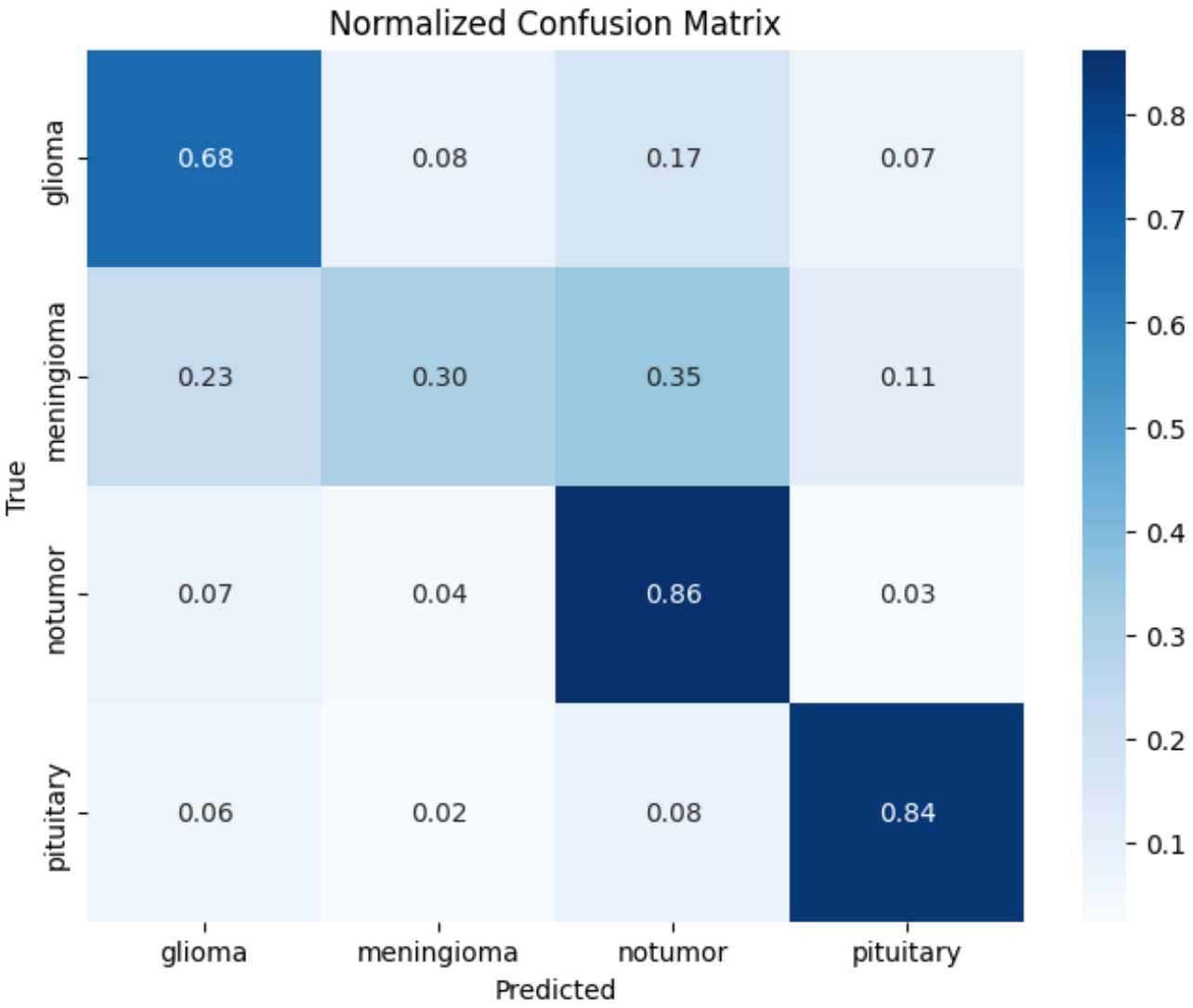

*RadImageNet DenseNet121*

**Figure III:** ***Per-class testing accuracy for ConvNeXt-Tiny, EfficientNetV2S, and RadImageNet DenseNet121.*** *ConvNeXt-Tiny (top-right) and EfficientNetV2S (top-left) both achieved perfect or near-perfect accuracy on the notumor and pituitary classes. However, RadImageNet DenseNet121 (bottom) exhibited markedly lower performance on meningioma (~26%), confirming its difficulty with this class as observed in the confusion matrices. It also underperformed on glioma and pituitary compared to the other models.*

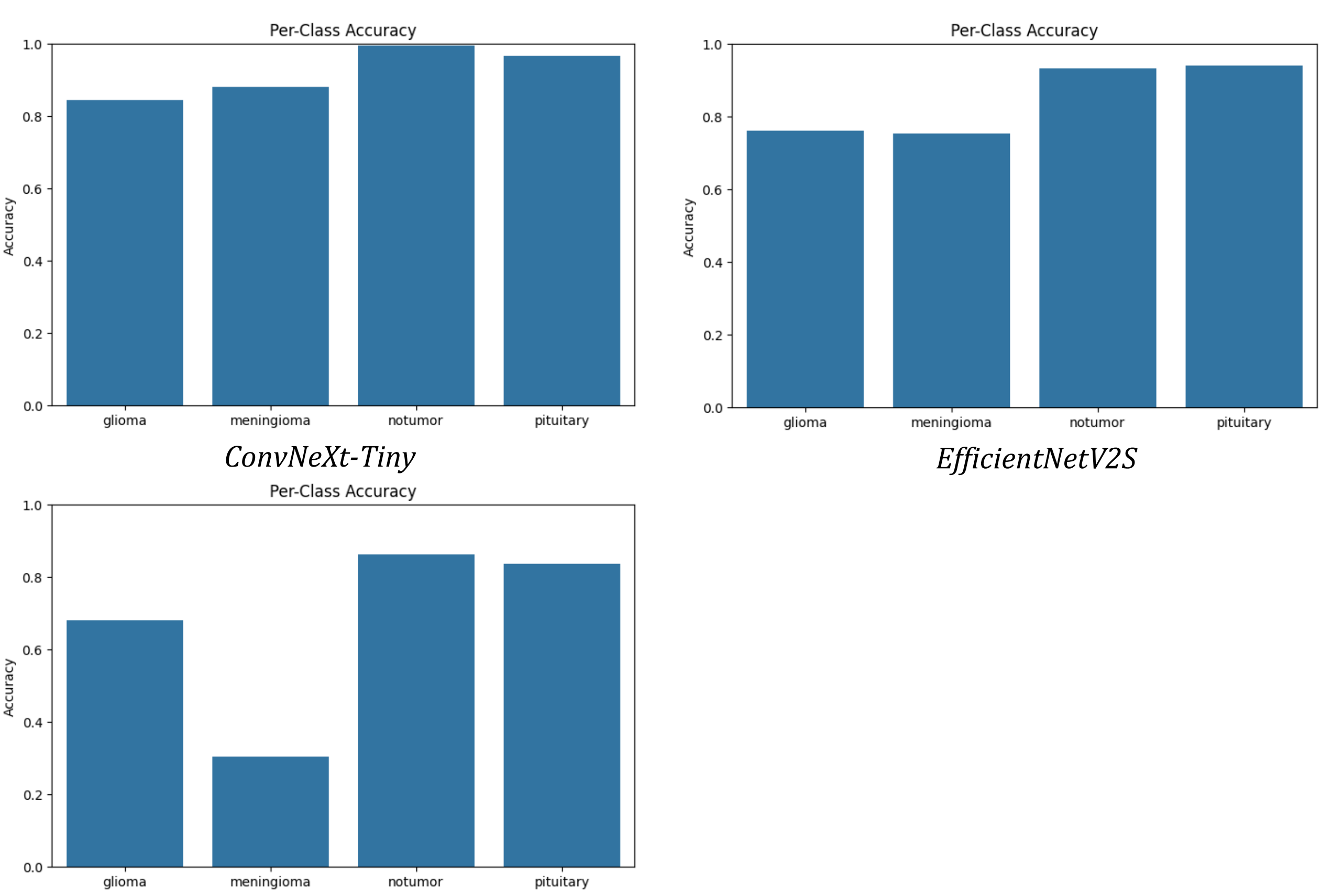

*ConvNeXt-Tiny*

*EfficientNetV2S*

*RadImageNet DenseNet121*

### 4.3 ROC and AUC

The ROC curves and AUC scores for each model are shown in Figure IV. ConvNeXt-Tiny outperformed the other models with a consistently higher true positive rate (TPR) and a larger area under the curve (AUC). The AUC scores for ConvNeXt-Tiny, EfficientNetV2S, and DenseNet121 were 98.5%, 96%, and 88%, respectively. The results clearly indicate that ConvNeXt-Tiny demonstrated superior discriminatory power in detecting brain tumor classes, followed by EfficientNetV2S and DenseNet121.

**Figure IV:** ***ROC curves and AUC scores for each model.*** *ConvNeXt-Tiny (top-right) achieved the highest overall discriminatory performance with a mean AUC of 98.5%, demonstrating excellent true positive rates across all classes. EfficientNetV2S (top-left) followed with a mean AUC of 96%, while RadImageNet DenseNet121 (bottom) showed comparatively lower performance with a mean AUC of 88%, consistent with its higher misclassification rates observed in Figures II and III.*

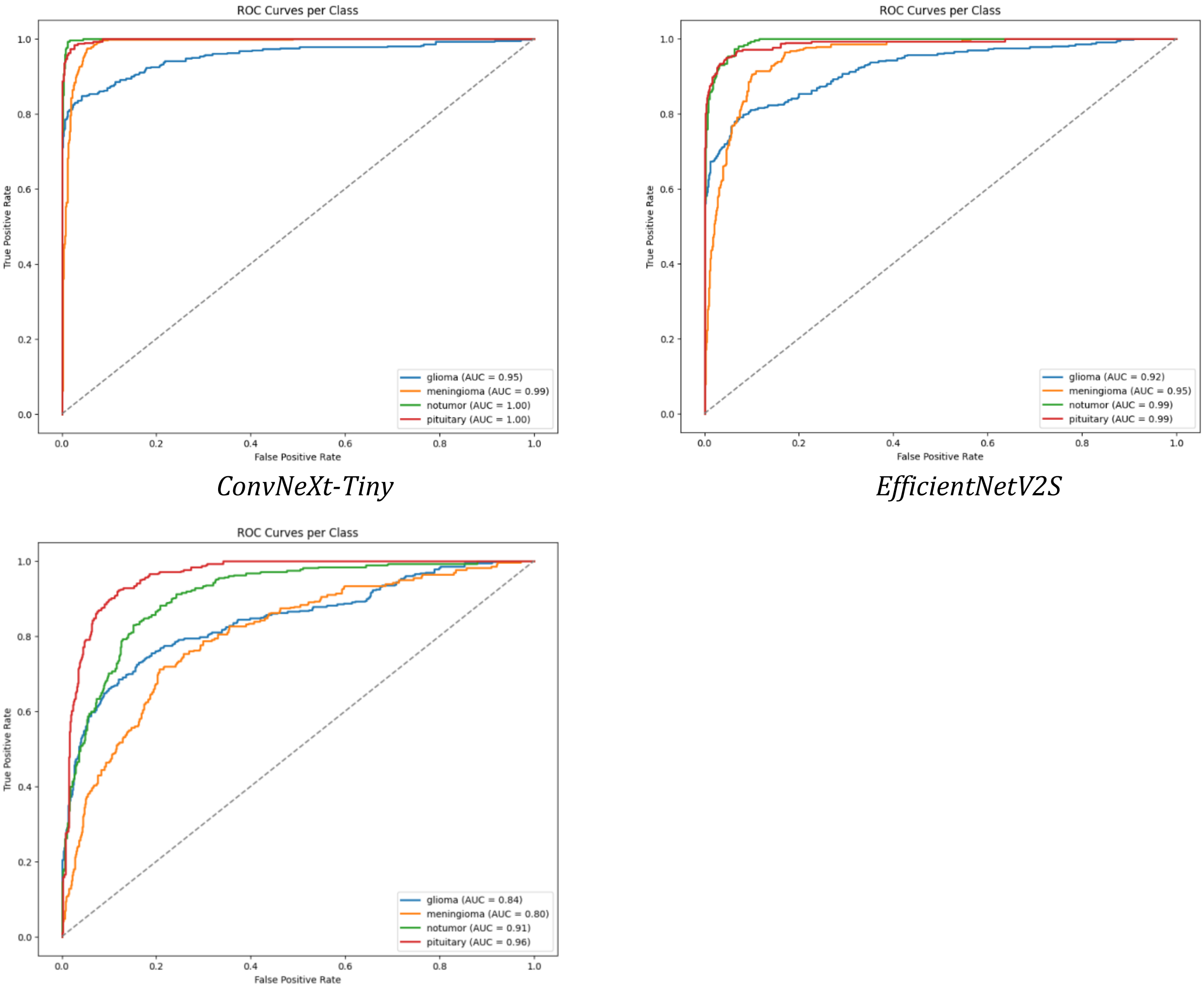

*ConvNeXt-Tiny*

*EfficientNetV2S*

*RadImageNet DenseNet121*

### 4.4 Precision-Recall Curves

Precision-recall curves, as shown in Figure V, further reinforce the findings from the ROC analysis. ConvNeXt-Tiny had the highest precision across all classes, followed by EfficientNetV2S, with DenseNet121 showing suboptimal performance, particularly in low-recall situations.

**Figure V:** ***PR curves for all evaluated models.*** *In alignment with the ROC analysis, ConvNeXt-Tiny (top-right) demonstrates superior precision-recall trade-off across all tumor classes, maintaining high precision even as recall increases. EfficientNetV2S (top-left) follows with moderately strong performance, while RadImageNet DenseNet121 (bottom) exhibits notably lower*

*precision, particularly in low-recall regions, confirming its reduced reliability in positive class identification under imbalanced conditions.*

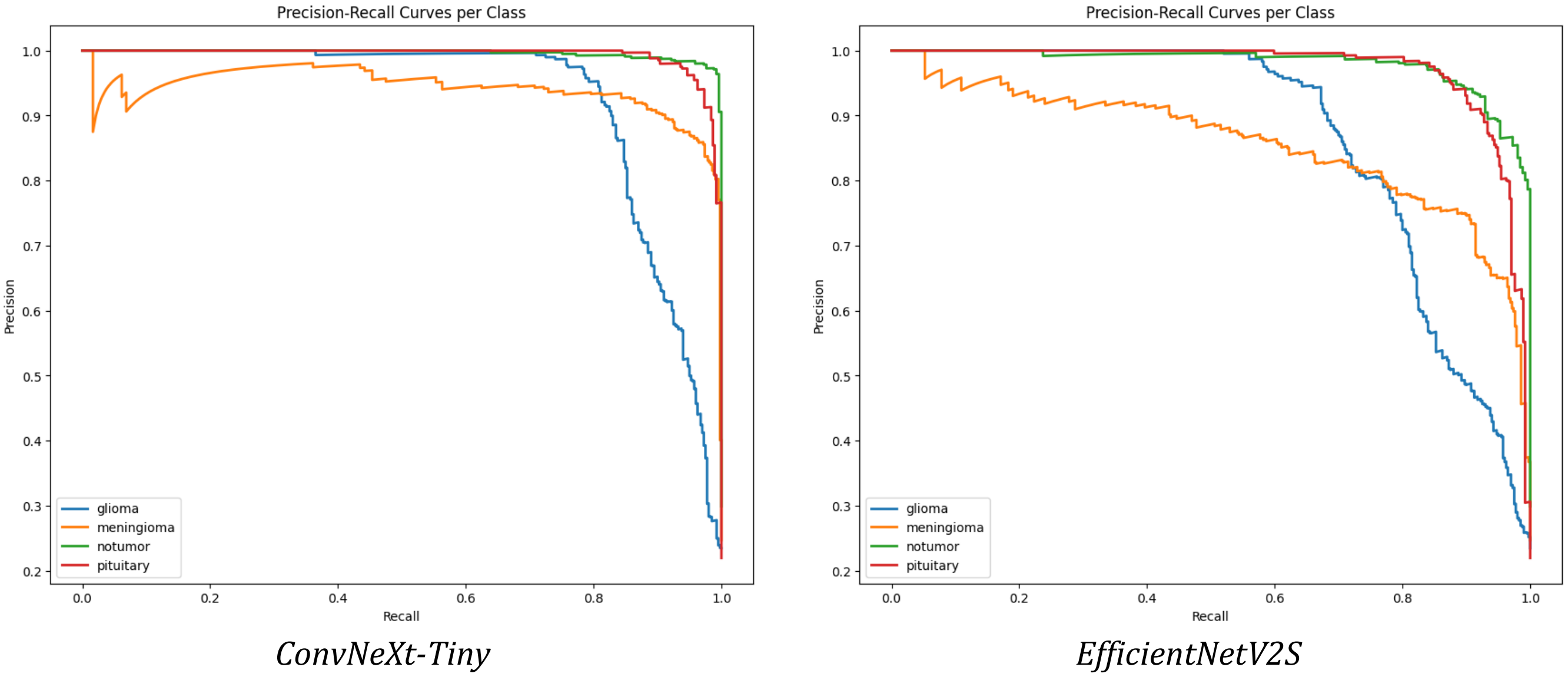

*ConvNeXt-Tiny* *EfficientNetV2S*

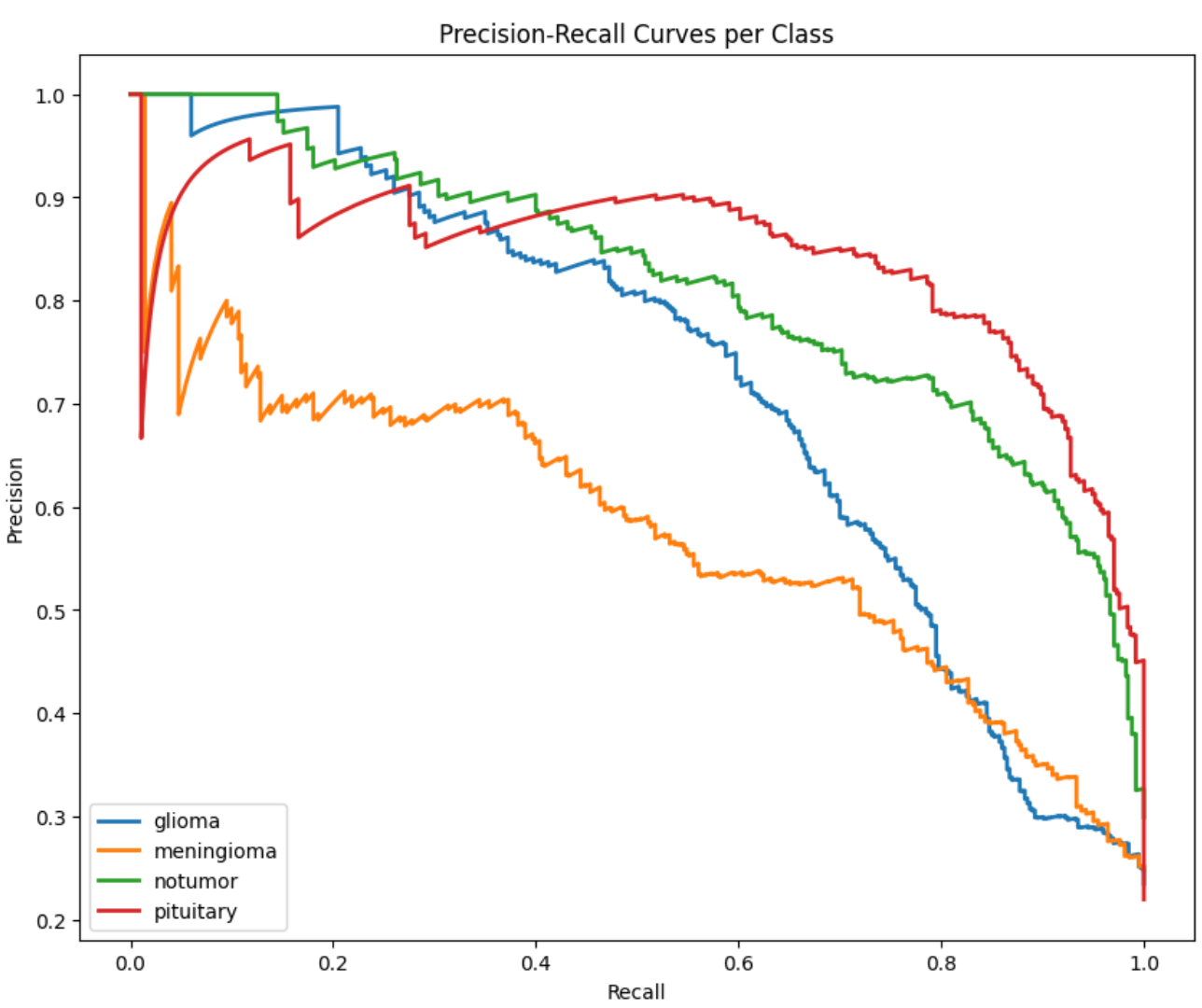

*RadImageNet DenseNet121*

## 5. Discussion

This study investigates the performance of three pretrained models in the task of brain tumor classification using MR images. Our findings indicate that general-purpose pretrained models, such as ConvNeXt-Tiny and EfficientNetV2S, perform better than domain-specific models like RadImageNet DenseNet121 when a small dataset is available. Among all the considered architectures, ConvNeXt-Tiny shows the best results on testing accuracy, AUC, and other metrics. This confirms that the models pretrained on large-scale general data may result in robust performance in medical imaging even with limited, domain-specific data.

## 6. Comparison of Model Performance

ConvNeXt-Tiny's superior performance is consistent with recent literature, which suggests that deeper architectures, which are pretrained on more diverse data, have a greater capacity for generalization [14]. DenseNet121, on the other hand, was domain-specifically pre-trained on medical images and underperformed substantially in this work, especially on smaller datasets. Such results may indicate that, while domain-specific pretraining can indeed be very helpful, it is by no means a guarantee of superior performance when datasets are limited. In fact, the obtained results align with studies that highlight the role of data scale, where general-purpose models trained on a variety of images can provide robust feature extraction capabilities that domain-specific models struggle to match under small-data conditions.

EfficientNetV2S performed moderately well, falling between ConvNeXt-Tiny and DenseNet121 regarding accuracy. While EfficientNet models are known for their efficiency and ability to perform well across a wide variety of tasks, ConvNeXt-Tiny's deeper architecture, specifically designed to handle a broader range of inputs, proved to be more beneficial for this task.

**7. Data Availability and Impact on Results**

One major limitation of this study is the availability of a limited dataset. Despite our efforts to augment the data and apply some preprocessing techniques, the small size of the dataset probably limited the generalization potential of the models, especially the domain-specific DenseNet121. The results of the experiments show that the DenseNet model would require more data to fully leverage its domain-specific pretraining and improve its performance. The scarcity of large, publicly available MRI datasets with tumor annotations presents a challenge for future research in this field. If additional similar structured datasets were available, it could allow for a more complete assessment and possibly help refine the conclusions obtained for this study.

**8. Generalization Across Different Datasets**

Our findings contribute to the ongoing debate about whether models pretrained on domain-specific data can outperform those pretrained on large-scale, general datasets. While domain-specific pretrained models have achieved much success when large datasets are available, our results indicate that, for small dataset sizes, general-purpose models perform better since they can be generalized against different types of data. This implies that both model architecture and dataset size become relevant in choosing an appropriate pretrained model for medical imaging applications.

**9. Future Directions**

In future work, gathering larger and more diverse brain tumor MRI datasets, ideally with a variety of tumor types, imaging protocols, and patient demographics, is desired. Additionally, incorporating explainability approaches, such as Grad-CAM and other visualization techniques, is crucial for increasing clinical practicality and building trust in model predictions. Hybrid models that combine domain-specific and general-purpose pretraining could also offer promising results.

**10. Conclusion**

This study investigates three different pre-trained CNN architectures: ConvNeXtTiny, EfficientNetV2S, and RadImageNet DenseNet121, which were used for brain tumor multi-class classification from two small MRI datasets. The performances of general-purpose models, especially the ConvNeXt-Tiny model, outperformed the domain-specific DenseNet121 model based on testing accuracy and generalization performance, even with its limited dataset size. EfficientNetV2S also demonstrated a good performance but slightly lower compared to ConvNeXt-Tiny.

The poor performance by DenseNet121 may suggest that domain-specific pretraining does not always lead to better results in a situation whereby there is limited availability of data. This study was considerably limited by the dataset size, which likely affects the performance of the models. Further work should focus on the use of larger datasets and explore hybrid approaches that combine domain-specific with general pretraining as ways of improving performance in medical imaging tasks.

Overall, this study reveals that general pretrained models like ConvNeXt-Tiny may present more reliable performance for brain tumor classification, especially under conditions with limited data.

## 11. Code Availability

The source code developed for this study, along with instructions for installation and replication of our experiments, is publicly available on GitHub at: https://github.com/Abedini03/Domain-Specific-vs-General-CNNs-Brain-MRI-Classification.git

## 12. References

[1] Dinggang Shen, Guorong Wu, and H.-I. Suk, "Deep Learning in Medical Image Analysis," 2017, doi: https://doi.org/10.1146/annurev-bioeng-071516-044442.

[2] Alexander Selvikvåg Lundervold and A. Lundervold, "An overview of deep learning in medical imaging focusing on MRI," 2019, doi: https://doi.org/10.1016/j.zemedi.2018.11.002.

[3] Geert Litjens *et al.*, "A survey on deep learning in medical image analysis," 2017, doi: https://doi.org/10.1016/j.media.2017.07.005.

[4] Puneet Tiwar, Jainy Sachdeva, Chirag Kamal Ahuja, and N. Khandelwal, "Computer Aided Diagnosis System-A Decision Support System for Clinical Diagnosis of Brain Tumours," *International Journal of Computational Intelligence Systems,* 2017, doi: https://doi.org/10.2991/ijcis.2017.10.1.8.

[5] Ibomoiye Domor Mienye, Theo G. Swart, George Obaido, Matt Jordan, and P. Ilono, "Deep Convolutional Neural Networks in Medical Image Analysis: A Review," 2025, doi: https://doi.org/10.3390/info16030195.

[6] Hee E. Kim, Alejandro Cosa-Linan, Nandhini Santhanam, Mahboubeh Jannesari, Mate E. Maros, and T. Ganslandt, "Transfer learning for medical image classification: a literature review," 2022, doi: https://doi.org/10.1186/s12880-022-00793-7.

[7] Xueyan Mei *et al.*, "RadImageNet: An Open Radiologic Deep Learning Research Dataset for Effective Transfer Learning," 2022, doi: https://doi.org/10.1148/ryai.210315.

[8] Harrison C. Gottlich *et al.*, "Effect of Dataset Size and Medical Image Modality on Convolutional Neural Network Model Performance for Automated Segmentation: A CT and MR Renal Tumor Imaging Study," 2023, doi: https://doi.org/10.1007/s10278-023-00804-1.

[9] M. Nickparvar. *Brain Tumor MRI Dataset*, doi: https://doi.org/10.34740/kaggle/dsv/2645886.

[10] S. Bhuvaji, A. Kadam, P. Bhumkar, S. Dedge, and S. Kanchan. *Brain Tumor Classification (MRI)*, doi: https://doi.org/10.34740/kaggle/dsv/12745533.

[11] X. Mei *et al.*, "RadImageNet: An Open Radiologic Deep Learning Research Dataset for Effective Transfer Learning," *Radiology: Artificial Intelligence,* 2022, doi: https://doi.org/10.1148/ryai.210315.

[12] M. Tan and Q. V. Le, "EfficientNetV2: Smaller Models and Faster Training," presented at the Proceedings of the 38th International Conference on Machine Learning, PMLR, 2021. [Online]. Available: https://arxiv.org/abs/2104.00298#.

[13] Z. Liu, H. Mao, C.-Y. Wu, C. Feichtenhofer, T. Darrell, and S. Xie, "A ConvNet for the 2020s," presented at the Proceedings of the IEEE/CVF Conference on Computer Vision and Pattern Recognition, 2022. [Online]. Available: https://ieeexplore.ieee.org/document/9879745.

[14] D. Hendrycks, X. Liu, E. Wallace, A. Dziedzic, R. Krishnan, and D. Song, "Pretrained Transformers Improve Out-of-Distribution Robustness," 2020, doi: https://doi.org/10.48550/arXiv.2004.06100.